\documentclass[letterpaper, 10 pt, conference]{ieeeconf}  

\IEEEoverridecommandlockouts                              

\usepackage{graphicx} 
\usepackage{amsmath} 
\usepackage{bbold}
\usepackage{color}
\usepackage{subcaption}
\usepackage[nice]{nicefrac}
\usepackage{hyperref}
\usepackage[colorinlistoftodos]{todonotes}
\usepackage{verbatim}
\usepackage{mathtools}
\usepackage{cite}
\usepackage{xspace}

\title{\huge
Indoor Localization for an Autonomous Model Car:
\\
A Marker-Based Multi-Sensor Fusion Framework 
}

\author{Xibo Li$^{*}$, Shruti Patel$^{*}$,  David Stronzek-Pfeifer$^{*}$ and Christof B\"uskens$^{*}$
	\thanks{$^{*}$The authors are with the working group Optimization and Optimal Control, Center for Industrial Mathematics, University of Bremen, 28359 Bremen, Germany
		{\small(email: \{lixibo, spatel, stronzek, bueskens\}@uni-bremen.de)}}%
	\thanks{\textit{Corresponding author: Xibo Li}}%
	\thanks{The work of Shruti Patel and David Stronzek-Pfeifer is supported by funds of the German Government’s Special Purpose Fund held at Landwirtschaftliche Rentenbank.
		The work of Xibo Li is supported by the Federal Ministry of Economic Affairs and Climate Action on the basis of a decision by the German Bundestag.
	}%
}

\begin{document}

\maketitle
\thispagestyle{empty}
\pagestyle{empty}

\begin{abstract}
Global navigation satellite systems readily provide accurate position information when localizing a robot outdoors. However, an analogous standard solution does not exist yet for mobile robots operating indoors.
This paper presents an integrated framework for indoor localization and experimental validation of an autonomous driving system based on an advanced driver-assistance system (ADAS) model car.
The global pose of the model car is obtained by fusing information from fiducial markers, inertial sensors and wheel odometry. In order to achieve robust localization, we investigate and compare two extensions to the Extended Kalman Filter; first with adaptive noise tuning and second with $\chi^2-$testing for measurement outlier detection. An efficient and low-cost ground truth measurement method using a single LiDAR sensor is also proposed to validate the results. The performance of the localization algorithms is tested on a complete autonomous driving system with trajectory planning and model predictive control. 
\end{abstract}


\section{INTRODUCTION}\label{sec:introduction}
Mobile autonomous robots in indoor environments are being deployed in a broad variety of applications, ranging from industrial facilities, hospitals and healthcare, to personal assistance robots at home and in the hospitality sector. 
Small indoor robots are also a useful tool in prototyping larger autonomous robots that function outdoors, such as self-driving cars. In this paper, we provide a sensor fusion framework for an ADAS model car, which serves as a comprehensive prototyping platform for autonomous driving algorithms before they are tested in the real world.
While the use of global navigation satellite systems (GNSS) like GPS is the standard method for determining pose of cars driving on the road, it is not suitable for the miniature model being tested indoors, necessitating the development of a distinct localization solution. Such a solution is essential for the rest of the autonomous driving pipeline, since motion planning, control and tactical decision making all depend on a precise pose estimate of the ego-vehicle.
\begin{figure}
	\centering
	\includegraphics[width=1\linewidth]{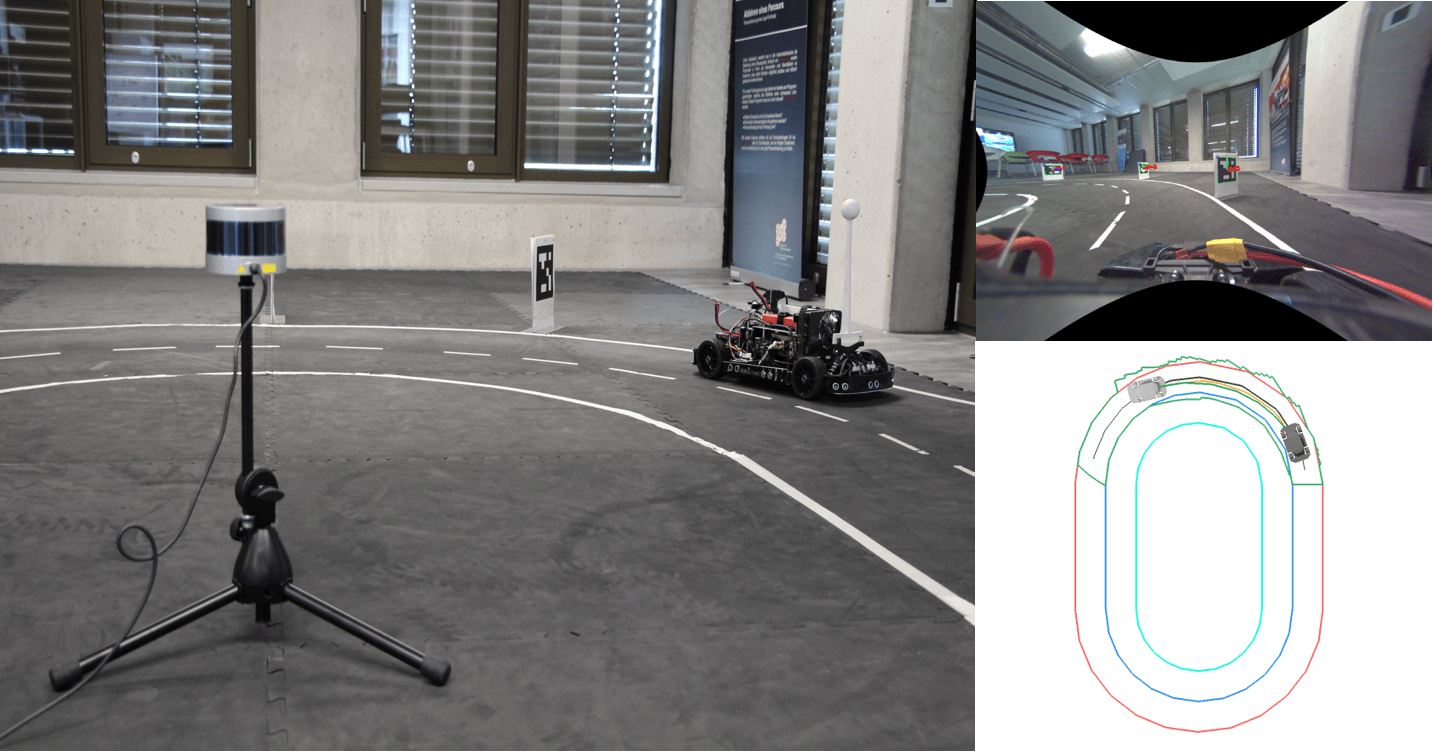}
	\caption{Clockwise from the left: the LiDAR, ADAS model car and ArUco marker in the first picture, view from the onboard camera in the second picture, and the trajectory planning map in the simulation in the third picture.}
	\label{fig:overview}
\end{figure}

Vision-based localization systems that determine robot pose indoors using natural or artificial landmark recognition in the robot's vicinity provide a popular low-cost alternative to technologies such as  WiFi, Bluetooth, ultra-wide band or radio frequency identification~\cite{XuLandmarks}. Simultaneous localization and mapping (SLAM) methods are suitable when the robot environment has sufficiently many feature points, the map of the environment is not known apriori, and where localization in a fixed global map is not necessary~\cite{9440682}\cite{6106777}. In sparse environments and in applications where the robot pose needs to be determined in a global map, artificial landmarks in the form of fiducial markers such as ArUco markers~\cite{garrido2014automatic} or AprilTags~\cite{5979561} are used. However, landmark-based techniques may lack precision due to sporadic measurements, sensitivity to light conditions, issues with re-initialization, and type of landmark. Meanwhile, methods that use motion information from wheel odometry and inertial sensors have more reliable measurements but fail to provide global positioning and accumulate integration errors\cite{Cho2011}. To address these limitations, a viable method is to use sensor fusion algorithms to combine position information from vision-based localization with inertial sensors and wheel odometry for a near-optimal state estimate~\cite{8066377}.


The Extended Kalman Filter (EKF) is often the conventional choice for non-linear systems due to its efficient use of computational resources and quick prototyping~\cite{kayhani2019improved}\cite{7868190}, however it is not guaranteed to work straightaway, and has several potential reasons for failing to provide a stable state estimate. Alternate algorithms such as the Unscented Kalman Filter or the Particle Filter can quickly become computationally expensive or time-intensive to implement. Several modifications to the Extended Kalman Filter have been proposed to rectify the faults that may cause it to perform poorly~\cite{SONG198659}\cite{8273755}\cite{1104658}, which makes understanding the underlying cause that affects the EKF performance essential to the final choice of a suitable algorithm.

In this paper we present a robust, cost-effective, and easy to implement EKF-based multi-sensor fusion methodology for global indoor positioning of an autonomous car prototype using ArUco markers, inertial measurement units (IMU) and wheel odometry. Moreover, we discuss in detail the methodology for obtaining the ground truth for the robot position using a LiDAR tracking system, as well as the calculations to extract pose measurements using an onboard camera that detects the ArUco markers. We investigate two modifications to the classic EKF algorithm, the so-called Adaptive Extended Kalman Filter (AEKF)~\cite{8273755}, which can dynamically tune process and measurement noise matrices of the filter and therefore can rectify poor filter performance incurred due to incorrect noise modeling, and the EKF with $\chi^2-$testing~\cite{chen2021autonomous}, which corrects state estimation errors incurring due to measurement outliers.
The entire framework is implemented and tested on a fully integrated autonomous driving system in the form of an ADAS model car (Fig.~\ref{fig:overview}), which includes the final trajectory planning and control of the system. We perform an in-depth comparison of the filter performances based on positioning error and smoothness of steering controls obtained from the fused pose estimates. Consequently, we determine the most suitable filter based on the main source of error affecting the EKF output.

The rest of the paper is organized as follows: In Section~\ref{sec:related_works} we overview related literature. Sections~\ref{sec:Sensor_Setup}, \ref{sec:scheme}, \ref{sec:ground_truth} and \ref{sec:aruco} describe the hardware platform and sensor setup, the scheme design of the software modules, the ground truth measurement and the ArUco marker observation, respectively.  We detail the sensor fusion algorithms used in this work in Section~\ref{sec:kf_algo}, and experimental setup and results are discussed in Section~\ref{sec:results}. Finally, Section~\ref{sec:conclusion} concludes the paper. A short video about the localization performance is available at \url{https://youtu.be/KpZkN8eUkb4}.

\section{RELATED WORKS \label{sec:related_works}}

There are two categories of works that deal with autonomous indoor localization using landmarks: one for unknown environments without prior maps, and the other for systems with pre-built maps. For robots operating in unstructured environments, SLAM is needed. Visual SLAM methods such as the ORB-SLAM technique presented in~\cite{7219438} and \cite{7946260} use cameras to extract feature points from a robot's surroundings and can be easily used in new environments to build up a map, and in~\cite{9440682} the authors additionally incorporate inertial sensors into the system. In~\cite{MUNOZSALINAS2020107193}, the authors fuse keypoints from natural landmarks with information encoded in ArUco markers and show that a combination of both works better than using each of them independently, whereas \cite{MUNOZSALINAS2019156} uses ArUco makers as SLAM keypoints and builds a map of planar markers. LiDAR SLAM methods such as one presented in~\cite{6106777} have been shown to combine onboard 2D mapping and a 6-DOF pose estimation while consuming low computational resources. 
An initial investigation was done for localizing our ADAS model car with SLAM methods. The ORB-SLAM2~\cite{7946260} and UcoSLAM techniques~\cite{MUNOZSALINAS2020107193} require a sufficiently high-resolution camera, in the absence of which feature-point extraction proves to be challenging. Consequently, we found that during sharp steering maneuvers, the algorithms tended to lose pose information and re-initialization of the model car was not robust with our low resolution camera. Moreover, these methods were not suitable in our setup with relatively sparse surroundings. Camera resolution was not an issue when using the LiDAR-based Hector SLAM method~\cite{6106777}, however, loss of pose occurred during the same sharp steering maneuvers, which lead to an unsuccessful alignment of the laser scan with the map. 

The second category of methods deals with fusing information from fiducial markers with data from inertial sensors and wheel encoders. These methods are suitable when simultaneous mapping is not required and the fiducial markers can be easily assigned position information, and have the additional benefit of having a highly reduced computation effort compared to SLAM. In~\cite{8066377}, an extended $H_{\infty}$ filter is used to fuse odometry and gyroscope data with QR-code landmark recognition for localizing a robotic walking assistant, using a SICK-S300 Expert laser scanner for the ground truth measurement.
Another application of using fiducial markers in the absence of GPS data is found in \cite{chen2021autonomous}, where the authors use April Tags and an EKF with $\chi^2$-testing for a consistent trajectory tracking of an unmanned underwater vehicle. However, the accuracy of the method tested in this work has not been validated with a ground truth measurement.  
Localization for an underground autonomous valet parking system is developed in~\cite{Fang2020MarkerBasedMA} using fiducial markers, odometry and feature points fused with a particle filter, and the ground truth estimate obtained by the LeGO-LOAM algorithm and laser scans.
While the technique discussed in the present paper shares conceptual similarities with the works described above, we place significantly more emphasis on developing and describing a comprehensive ground truth estimation system, without which a reliable validation of algorithms is not possible. Moreover, we choose to refine the EKF instead of opting for more complex filters, which results in a comparably easy and low-cost solution without compromising on robustness. Additionally, the result of the localized robot pose is input into a trajectory planning and control system, thereby substantiating the applicability of the method into a fully integrated autonomous driving framework. Finally, the methods outlined in this paper can be easily generalized to more complex indoor environments.

\section{METHODOLOGY \label{sec:method}}

\subsection{Sensor Setup of ADAS Model Car}
\label{sec:Sensor_Setup}
\begin{figure}
	\centering
	\includegraphics[width=0.8\linewidth]{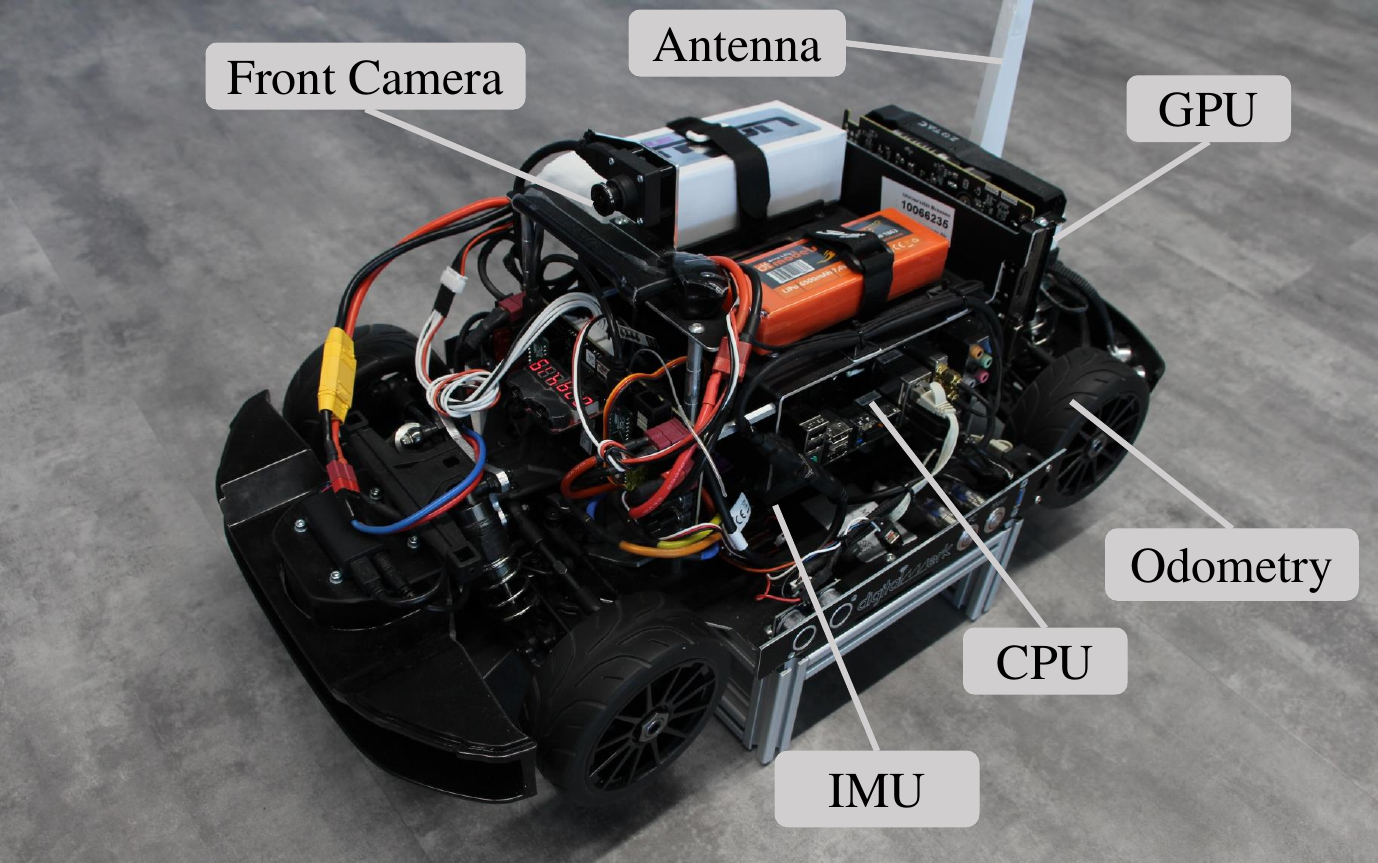}
	\caption{Hardware configuration of the model car}
	\label{fig:sensor_setup}
\end{figure}

Our indoor localization frame is implemented on the ADAS model car, a miniature vehicle platform developed by Digitalwerk~\cite{digitalwerk}. It is designed as a 1:8 replica of the Audi Q2 equipped, among other sensors, with an RGB camera and an IMU, and mimics the dynamic behavior of a real car driving on a road, making it an ideal platform for prototyping autonomous driving algorithms \cite{berger2014competition}. Fig.~\ref{fig:sensor_setup} shows the sensor setup used for localizing this model car indoors. The front camera with a resolution 1280$\times$960 px is used to detect the fiducial positioning markers, and measures the relative distance between the marker and the camera. The fiducial makers used for this purpose are the ArUco Markers~\cite{garrido2014automatic}. Meanwhile, the Invensense MPU-9250 motion tracking device is the IMU that measures the yaw rate of the vehicle. Two separate Honeywell HOA0902-11 Transmissive Encoder Sensors with a resolution up to 0.457~mm measure the speed of the two rear wheels. The dimensions of the model car are 610$\times$320~mm with a wheelbase of 360~mm. Its minimum turning radius is 0.6 m with the maximum forwards speed over 2 m/s. Details of additional hardware components of the ADAS model car system are listed in Table \ref{tab:hardware_information}.
\begin{table}
	\medskip
	\begin{center}
		\caption{Overview of Hardware Components}
		\label{tab:hardware_information}
		\begin{tabular}{c c}
			\hline
			Component & Item\\
			\hline	
		    Mainboard & GIGABYTE GA-Z170N-WIFI miniITX\\
		    CPU & Intel$\textregistered$  Core i3-6100T CPU (3.2 GHz)\\
	        GPU & NVIDIA GeForce GTX1050Ti\\
	        Memory & 128GB M.2 2280 SSD\\
	        RAM & 8 GB DDR4 PC-2133 RAM\\
	        Speed Motor & Hacker SKALAR 10 21.5 Brushless Motor\\
	        Steering Servo & Absima "ACS1615SG"\\
			\hline
		\end{tabular}
	\end{center}

\end{table}

\subsection{Scheme Design}
\label{sec:scheme}
\begin{figure}
	\medskip
	\centering
	\includegraphics[width=1\linewidth]{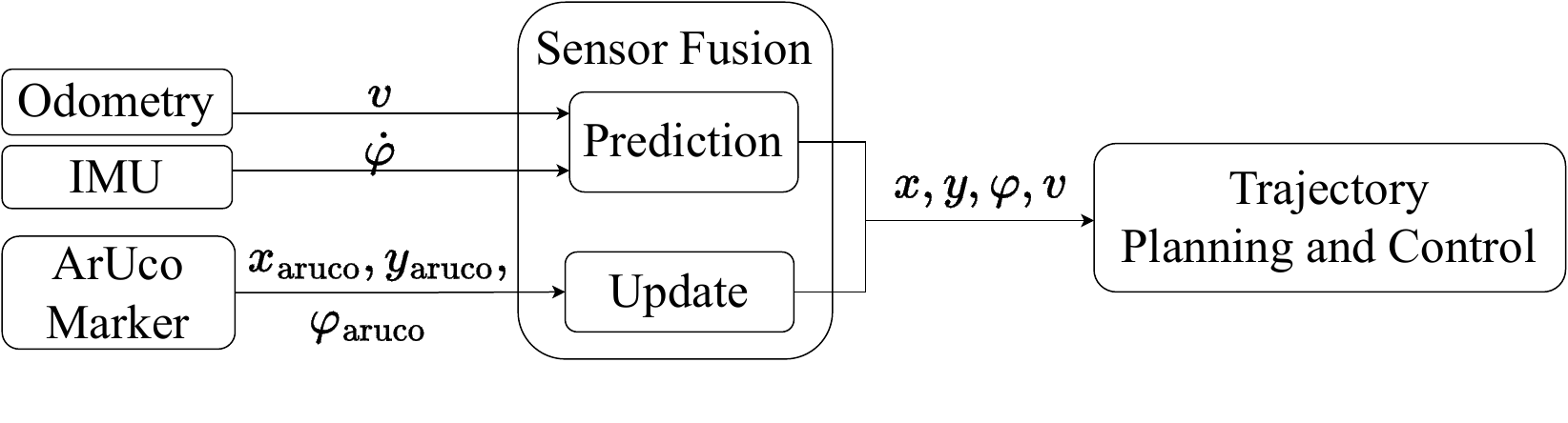}	
	\caption{Scheme design of the indoor autonomous driving framework.}
	\label{fig:flow_chart}
\end{figure}

The localization, trajectory planning and control algorithms are implemented on the hardware platform detailed in Sec.~\ref{sec:Sensor_Setup}. Fig.~\ref{fig:flow_chart} describes the complete indoor autonomous driving scheme, where the localization depends on three different sensors. As stated in the Section \ref{sec:Sensor_Setup}, the wheel speeds and the yaw rate are measured separately by the wheel encoder and IMU respectively, which provide the input for the prediction step of the sensor fusion algorithm used. The input for the update step comes from the position and yaw measurements provided by the ArUco markers, read by the RGB camera.  The subsequent fused state estimate output from our localization algorithms is then used together with a trajectory planning algorithm as input to a Model Predictive Controller (MPC) to generate control signals for the low-level controller~\cite{folkers2020time}\cite{rick2019autonomous}. 

\subsection{Ground Truth Measurement}
\label{sec:ground_truth}
The evaluation of the localization algorithms presented in this work requires a precise ground truth estimate. OptiTrack motion capture system is a common solution for that. However, a large indoor environment measuring more than 6$\times$8 m requires the mounting of many cameras, which leads to high costs.
A common low-cost approach for this is to use fixed laser scanners to track the robot motion and measure its position accurately \cite{Piardi2018DevelopmentOA}. 
Other approaches for indoor localization include installing a fixed camera, as presented in \cite{Ceriani2009}, where the authors also present a network of fixed laser scanners for ground truth estimates. 
In this work, we use a setup consisting of a single fixed laser scanner, which proves to be sufficient for tracking the model car in the existing room topography.

The ground truth tracking system consists of a Velodyne VLP-16 LiDAR with surround view as well as a simple reference object of known dimensions placed above the vehicle.
The LiDAR is placed at a height of approximately 0.46 m centrally with respect to the intended track of the model car to obtain the best possible coverage of the area of interest, as shown in Fig.~\ref{fig:overview}. An antenna with a spherical top is mounted at the center of the rear axle of the model car, serving as the reference object to be tracked by the LiDAR for a cleaner point cloud output.
\begin{figure}
	\centering
	\begin{subfigure}[t]{0.58\linewidth}
		\includegraphics[width=\linewidth]{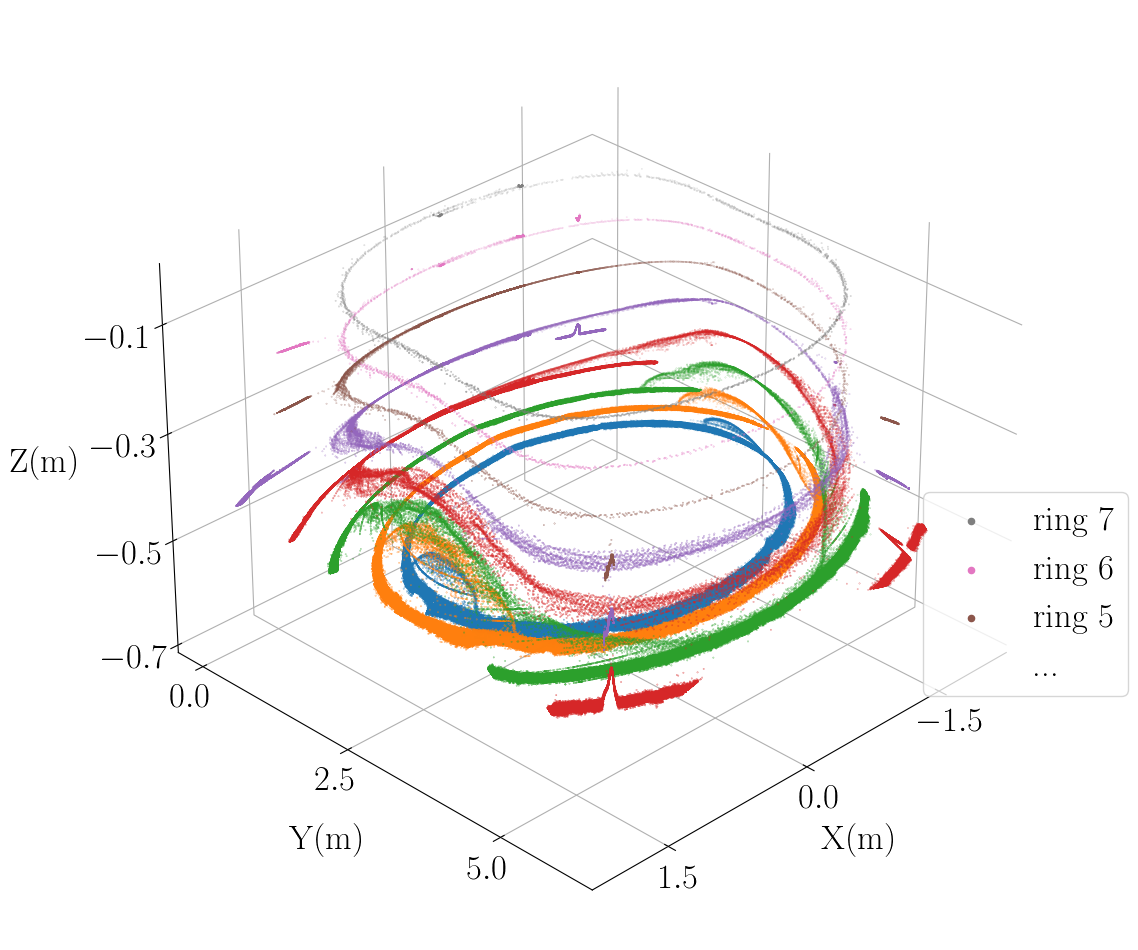}
		\caption{LiDAR point cloud from tracking the model car}
		\label{fig:plot_reference_trajectory_pointCloud}
	\end{subfigure}
	\begin{subfigure}[t]{0.4\linewidth}
		\centering
		\includegraphics[width=\linewidth]{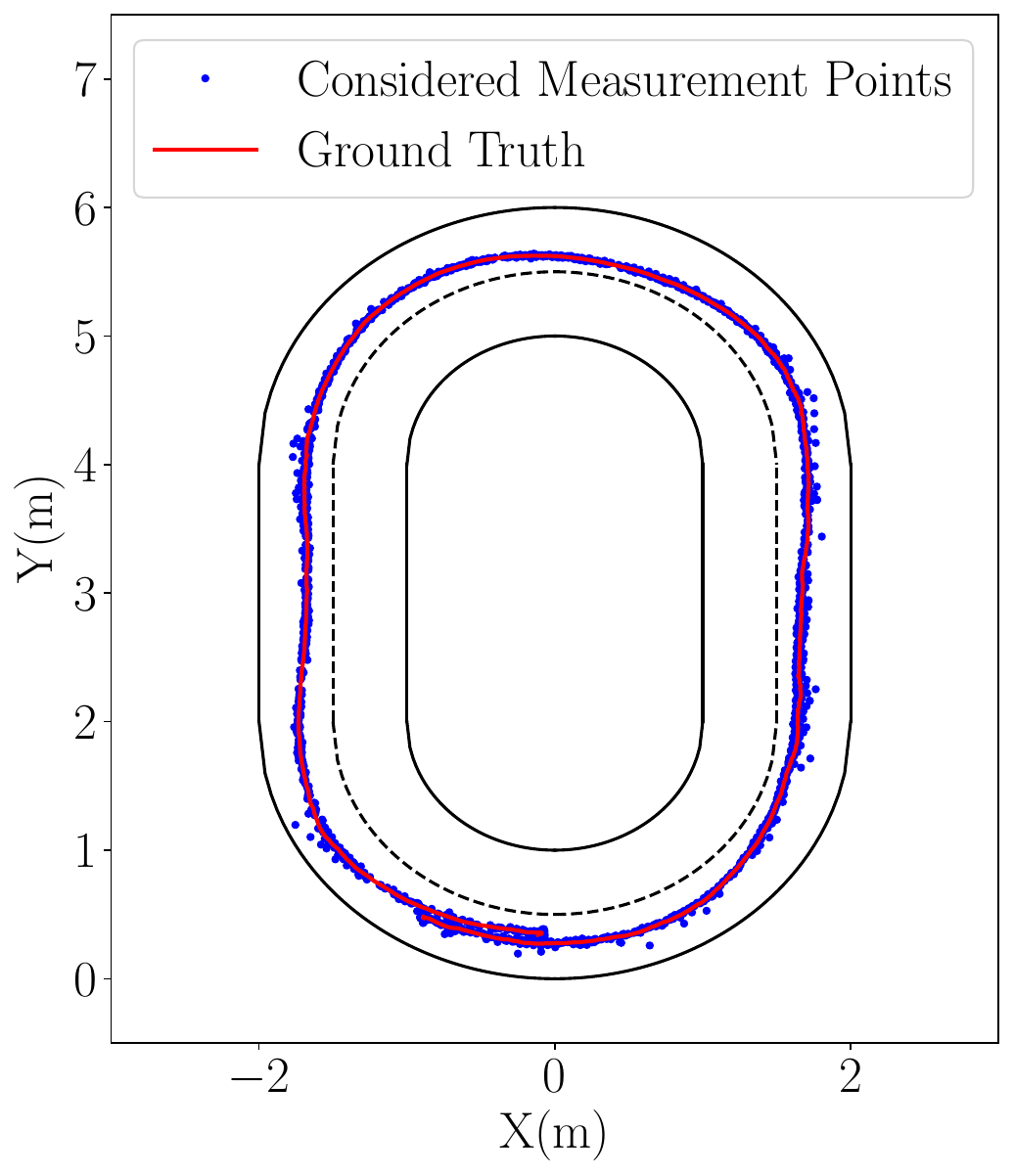}
		\caption{Reference trajectory}
		\label{fig:plot_reference_trajectory}
	\end{subfigure}
	\caption{(a) shows original 3D point cloud data with different color coding of the rings $0-7$.  In (b) an interpolated reference trajectory based on ring 7 is shown in 2D. }
\end{figure}
The surround scan of the LiDAR features 16 individual channels (rings) with different fixed vertical angles $\omega$ 
\cite{vlp-16}.
Fig. \ref{fig:plot_reference_trajectory_pointCloud} shows the eight lower vertical angles  $\omega$ from $-1^{\circ}$ to $-15^{\circ}$.
Within the controlled experimental setup, the partition of the point cloud data may be obtained by vertical ring angles since the height of the sphere and the laser scanner's position were chosen accordingly. It should be mentioned that during the experiment the ground on which the track ran was not absolutely flat. This results in a slight difference of height within the point cloud data as seen in Fig. \ref{fig:plot_reference_trajectory_pointCloud}.
Furthermore, the position coordinates of the sphere obtained from the LiDAR point cloud are corrected for its radius, since the laser scanner data only reflects off its surface. 
The position of each LiDAR measurement point is collected and the final smooth reference trajectory is calculated offline by using spline interpolation, as seen in Fig.~\ref{fig:plot_reference_trajectory}. The calculated trajectory is only used as the ground truth in order to compare with our localization algorithms. Note that this method can be generalized to more complex maps than the one shown here using multiple LiDARs to ensure coverage.

\subsection{ArUco Marker Measurements \label{sec:aruco}}
\begin{figure}
	\centering
	\includegraphics[width=0.9\linewidth]{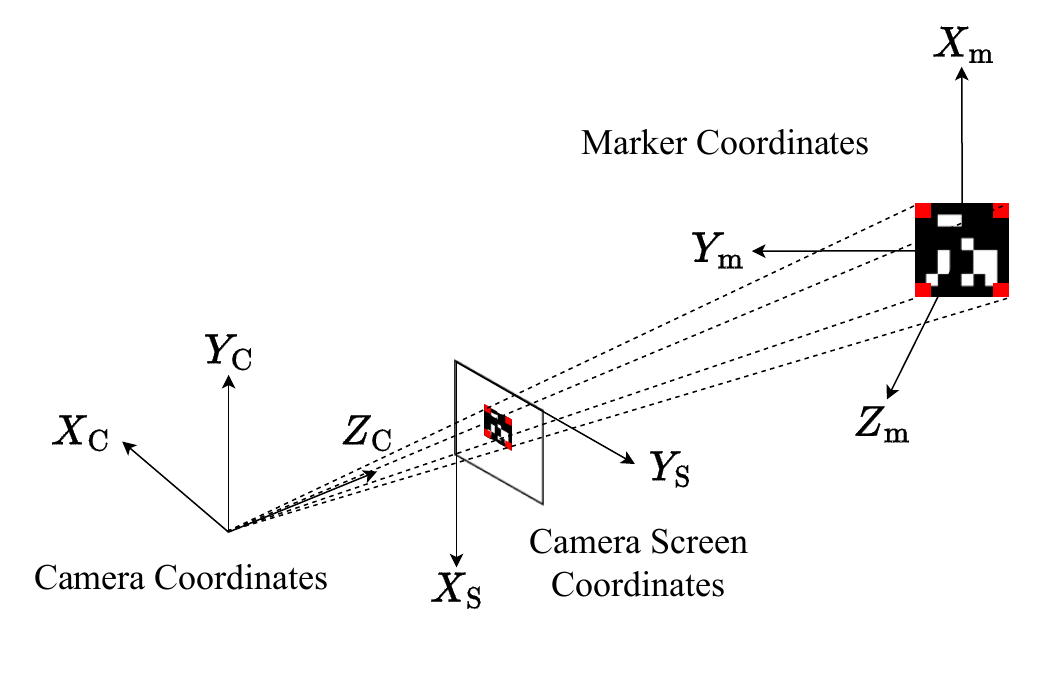}
	\caption{The relationship between marker coordinates and
		the camera coordinates is estimated by Perspective-n-Point (PnP) pose estimation.}
	\label{fig:aruco_marker_scheme}
\end{figure}

\begin{figure}[htb!]
	\centering
	\includegraphics[width=0.8\linewidth, trim = 1cm 0cm 0cm 0cm, clip]{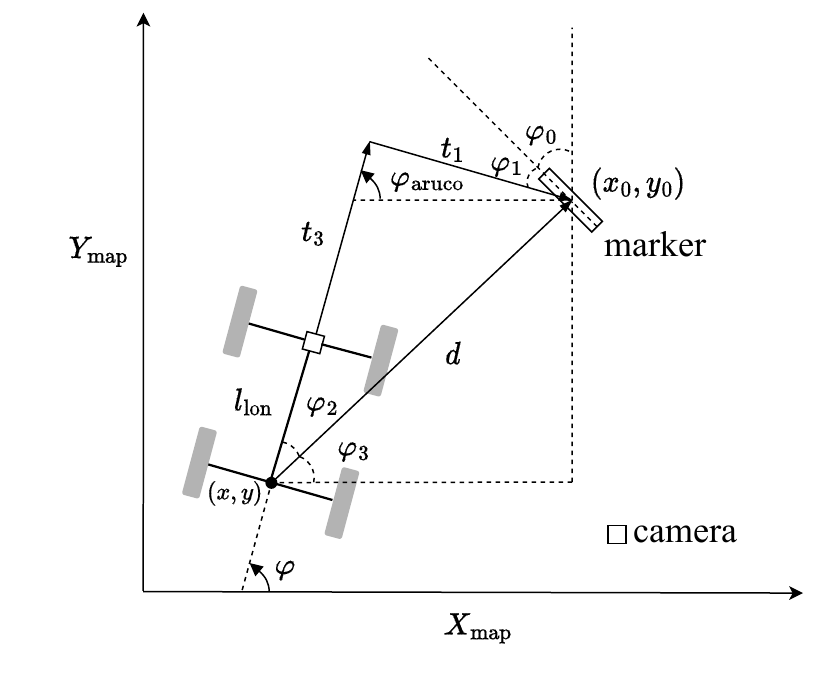}
	\caption{A schematic of the single track model and the geometry of pose detection via ArUco markers. The axes depict the global map frame.}
	\label{fig:single-track}
\end{figure}

For autonomous navigation in outdoor scenarios, positioning using GNSS augmented with real-time kinematic (RTK) corrections is the conventional solution.
However, satellite-based positioning is not a viable option when localizing indoor robots due to low signal strength and accuracy. Therefore, a robust and precise method for determining the global position of an indoor mobile robot such as our model car is necessary. 
As stated in Section \ref{sec:introduction}, there are varied approaches to solve the indoor robot localization problem. 
Here we propose a low-cost, efficient, and accurate method with multiple inputs. The well-known ArUco Marker positioning \cite{garrido2014automatic} is applied to estimate the relative pose between the visual marker and the camera mounted on the model car. 
The relative distance and heading between the camera and the marker are determined by detecting the marker corners (depicted by red dots in Fig.~\ref{fig:aruco_marker_scheme}) in the camera frame $\lbrace X_c, Y_c, Z_c \rbrace$, as well their 2D projection in pixels in the camera-screen frame $\lbrace X_s, Y_s\rbrace$.
The transformation between the four marker-corner coordinates  $(x_{ci}, y_{ci}, z_{ci})$ in the camera frame, and their 2D pixel points $( x_{si},y_{si})$ in the camera-screen frame is 
\begin{align}
	s\begin{bmatrix}
		x_{si}\\ 
		y_{si}\\ 
		1
	\end{bmatrix}= \begin{pmatrix}
		f_{x} & 0   & x_{s_0} \\ 
		0   & f_{y} & y_{s_0} \\ 
		0   & 0   & 1 
	\end{pmatrix}\begin{bmatrix}
		r_{11} & r_{12} & r_{13} & t_1\\ 
		r_{21} & r_{22} & r_{23} & t_2\\ 
		r_{31} & r_{32} & r_{33} & t_3
	\end{bmatrix}\begin{bmatrix}
		x_{ci}\\ 
		y_{ci}\\ 
		z_{ci}\\ 
		1
	\end{bmatrix}\,,
	\label{eq::aruco_marker_transition_matrix}
\end{align}
where the $s$ is a scaling factor,  $i \in \{1,\cdots,4\}$ is the four marker-corner points, $(x_{s_0},y_{s_0})$ are the coordinates of the principal point and $f_x, f_y$ are the scale factors in $X_s$ and $Y_s$ axes. The matrix of intrinsic parameters $x_{s_0},y_{s_0},f_x, f_y$ can be solved by a camera calibration procedure~\cite{opencv_aruco_calibration}. The $r_{ij}$ and $t_i$ are elements of the rotation matrix $R_c$ and the transition matrix $T_c$ between the camera-screen coordinates and camera coordinates. Determining the extrinsic matrices $R_c$ and $T_c$ is a general perspective-n-point camera pose determination (PnP) problem, which is solved by the OpenCV library solvePnP with different approaches~\cite{opencv_pnp} \cite{lepetit2009epnp}. 
After calculating the $R_c$ and $T_c$ matrices from Eq.~\ref{eq::aruco_marker_transition_matrix}, the relative angle $\varphi_1$ and distance $d$ between the camera and the ArUco marker can be estimated via 
\begin{align}
\varphi_1 = \arctan\left ( \frac{-r_{31}}{\sqrt{r_{32}^2+ r_{33}^2} } \right ),
\label{eq::aruco_marker_yaw_estimate}
\end{align}
\begin{align}
	d=\sqrt{(t_1+l_{\text{lat}})^2+(t_3+l_{\text{lon}})^2},
	\label{eq::aruco_marker_distance_estimate}
\end{align}
where the $l_{\text{lat}}$ and  $l_{\text{lon}}$ are the lateral and longitudinal distance in meters between the camera position and the center point of the rear axle of the vehicle. The geometrical relationships between the quantities described here is depicted in Fig.~\ref{fig:single-track}.
Each ArUco marker is assigned a pre-defined global position $(x_{0}, y_{0}, \varphi_{0})$ in the map frame.
The global position of the model car is calculated based on the relative distance $d$, the relative angle $\varphi_1$ and the absolute position of each ArUco marker.
Consequently, the heading angle of the model car in the global frame, $\varphi_{\text{aruco}}$, can be derived as
\begin{align}
\varphi_{\text{aruco}} = \varphi_{1} + \varphi_{0}.
\label{eq::aruco_marker_heading_estimate}
\end{align}
In our setup, the position of the camera is fixed to the center of the vehicle's front axle. Therefore the $l_{\text{lat}}$ is equal to 0. The angles $\varphi_{2}$ and $\varphi_{3}$ can be calculated as 
\begin{align}
\varphi_{2} &= \arctan\left ( \frac{t_{1}} {t_{3} + l_{\text{lon}}} \right )\,,\\
\varphi_{3} &= \varphi_{\text{aruco}} - \varphi_{2}\,,
\label{eq::aruco_marker_angle_correction}
\end{align}
which finally leads to the position of the model car in meters, measured in the map frame,
\begin{align}
x_{\text{aruco}} &= x_0 - d  \cos\left ( \varphi_{3}  \right ),\\
y_{\text{aruco}} &= y_0 - d  \sin\left ( \varphi_{3}  \right ).
\label{eq::aruco_marker_state_estimate}
\end{align}


\subsection{Indoor Localization Algorithms}
\label{sec:kf_algo}
In this paper we investigate the well-known non-linear variation of the Kalman Filter, the Extended Kalman Filter (EKF), with additional adaptive noise tuning~\cite{8273755} and $\chi^2-$testing \cite{1104658}\cite{chen2021autonomous} in order to achieve a precise state estimate of the system. The global state of the system is characterized by the vector $\mathbf{x} = [x, y, \varphi, v]^{\top}\,,$ where $(x, y)$~[m] are the Cartesian coordinates of the center of the rear-axle of the model car in the global map frame, $\varphi$~[deg] is the yaw angle with $\varphi = 0^{\circ}$ when the model car is headed along the $X_{\text{map}}$-axis and increases anti-clockwise. Finally, $v$ [m/s] is the vehicle speed. A schematic of the state variables can be seen in Fig.~\ref{fig:single-track}. 

At any time instance $t$, the state evolves via a non-linear system model which predicts $\mathbf{x}_{t|t-1}$ in terms of the previous state estimate $\mathbf{x}_{t-1}$, the control vector $\mathbf{u}_t$, and the additive process noise $\mathbf{w}_t$,
such that 
\begin{equation}
\mathbf{x}_{t|t-1} = f(\mathbf{x}_{t-1}, \mathbf{u}_t) + \mathbf{w}_t\,,
\end{equation}
with the process noise matrix $Q_t = E[\mathbf{w}_t\mathbf{w}_t^{\top}]$. 
The system model used in the subsequent filters is a kinematic single track model \cite{Schramm2014},
\begin{align}
f(\mathbf{x}_{t-1}, \mathbf{u}_t) = 
\begin{bmatrix}
	x_{t-1} + \Delta t  (v_{t-1} \cos \varphi_{t-1}) \\
 	y_{t-1} + \Delta t  (v_{t-1} \sin \varphi_{t-1}) \\
 	\varphi_{t-1} +  \Delta t  \omega_{t} \\
 	v_{t-1}	+ \Delta v
\end{bmatrix}
\label{eq:system-model}
\end{align}
where $\Delta t$ is the time-step. The yaw rate $\omega_t$ [deg/sec] is measured by the IMU, which forms the system's control input $\mathbf{u}_t = \omega_t$. The speed $v_{t-1} + \Delta v$ [m/sec] is directly obtained by the wheel encoders, instead of integrating the acceleration from the IMU, as the latter method is not robust during sharp turns. The EKF algorithm linearizes the system model through a first-order Taylor expansion around the fused state $\mathbf{x}_{t-1}$, which requires the calculation of the Jacobian $F_{t} = \frac{\partial f}{\partial \mathbf{x}}\bigg|_{\mathbf{\mathbf{x}_{t-1}}, \mathbf{u}_{t}}\,.$

While the calculation of the Jacobian can be computationally expensive in cases where it has to be calculated numerically, in the case of a single track model, an analytical expression for $F_t$ is trivially calculated from Eq.~\ref{eq:system-model}. 
The modeling of the process noise matrix $Q_t$ can be a very involved process, but an easy initial guess could be made by using the variances of $\omega_t$ determined from sensor measurements. In practice, however, $Q_t$ is essentially a hyperparameter which is tuned to enhance filter performance.

The measurement vector is ${\mathbf{z}_t = [x_{\text{aruco}}, y_{\text{aruco}}, \varphi_{\text{aruco}}]^\top}$, with measurements for the position and yaw made using the ArUco markers as described in Section \ref{sec:aruco}. 
Therefore the measurement model is linear and the relationship between the state and the measurements can be expressed as ${\mathbf{z}_t = H_t \cdot \mathbf{x}_t}$ with
$H_t = \left[ \text{I}_{3\times3} \quad 0_{3\times1}\right]$. The corresponding measurement noise matrix $R_t$ is set using the variances of the individual sensor measurements to $ R_t = \text{diag}\left(\sigma^2_{x},  \sigma^2_{y},  \sigma^2_{\varphi}\right)$.

Typically in EKF implementations a \textit{prediction} and a \textit{correction} step are computed at each $\Delta t$, however in our implementation the \textit{correction} step is carried out only when the measurements from the ArUco makers are available through the camera.
The filter performance is known to be highly sensitive to the choice of process and measurement noise matrices $Q_t$ and $R_t$ \cite{Mohamed1999AdaptiveKF}, which can be treated as hyperparameters of the algorithm, as well as to measurement outliers, in this case from the ArUco markers. In order to investigate the effects of these two aspects on filter performance, we enhance the EKF algorithm in two ways. The first algorithm, called the AEKF, uses an \textit{innovation}-based method to adaptively adjust $Q_t$ at each time-step, and a \textit{residual}-based method to adaptively adjust $R_t$. In the second algorithm, we investigate an EKF with $\chi^2-$testing of innovations to remove outlier measurement points.

\subsubsection{Adaptive Extended Kalman Filter}
Following the approach in \cite{Mohamed1999AdaptiveKF}, we compute the measurement \textit{innovation}
\begin{align}
\mathbf{d}_t = \mathbf{z_t} - H_t \mathbf{x}_{t|t-1},
\end{align}  
where $\mathbf{x}_{t|t-1}$ is the priori state estimate determined via Eq.~\ref{eq:system-model}, and the \textit{residual} 
\begin{align}
\mathbf{\varepsilon}_t = \mathbf{z}_t - H_t \mathbf{x}_t,
\end{align}
where $\mathbf{x}_t$ is the corrected state estimate at time $t$. The process and measurement noise matrices can therefore be adaptively calculated as
\begin{align}
Q_t =&\, \alpha_Q Q_{t-1} + (1 - \alpha_Q)(K_t \mathbf{d}_t \mathbf{d}_t^{\top} K_t)\,, \\
R_t =&\, \alpha_R R_{t-1} + (1 - \alpha_R)(\mathbf{\varepsilon}_t \mathbf{\varepsilon}_t^{\top} + H_t P_{t|t-1} H_t^{\top})\,,
\end{align}
where $K_t$ is the Kalman Gain, $P_{t|t-1}$ is the priori state covariance matrix corresponding to $\mathbf{x}_{t|t-1}$, and ${0 < \alpha_{Q,R} \leq 1}$ are the \textit{forgetting factors}. Setting $\alpha_{Q,R} = 1$ results in the non-adaptive, classic EKF. The schematic for the AEKF is shown in Fig.~\ref{fig:aekf-chisq}.

\subsubsection{Extended Kalman Filter with $\chi^2$-testing}

The $\chi^2-$~testing of innovations is used to detect faults in measurements and filter out outliers that would otherwise cause the localization performance to deteriorate. In order to do that we define the quantity $\chi^2$ as \cite{1104658}\cite{chen2021autonomous}
\begin{align}
\chi^2 =& \, \mathbf{d}_t^{\top}S_t^{-1}\mathbf{d}_t\,,\\
\text{where} \qquad S_t =& \, H_t P_{t|t-1}H_t^{\top} + R_t.
\end{align}   
\begin{figure}[htb!]
	\centering
	\medskip
	\includegraphics[width=\linewidth]{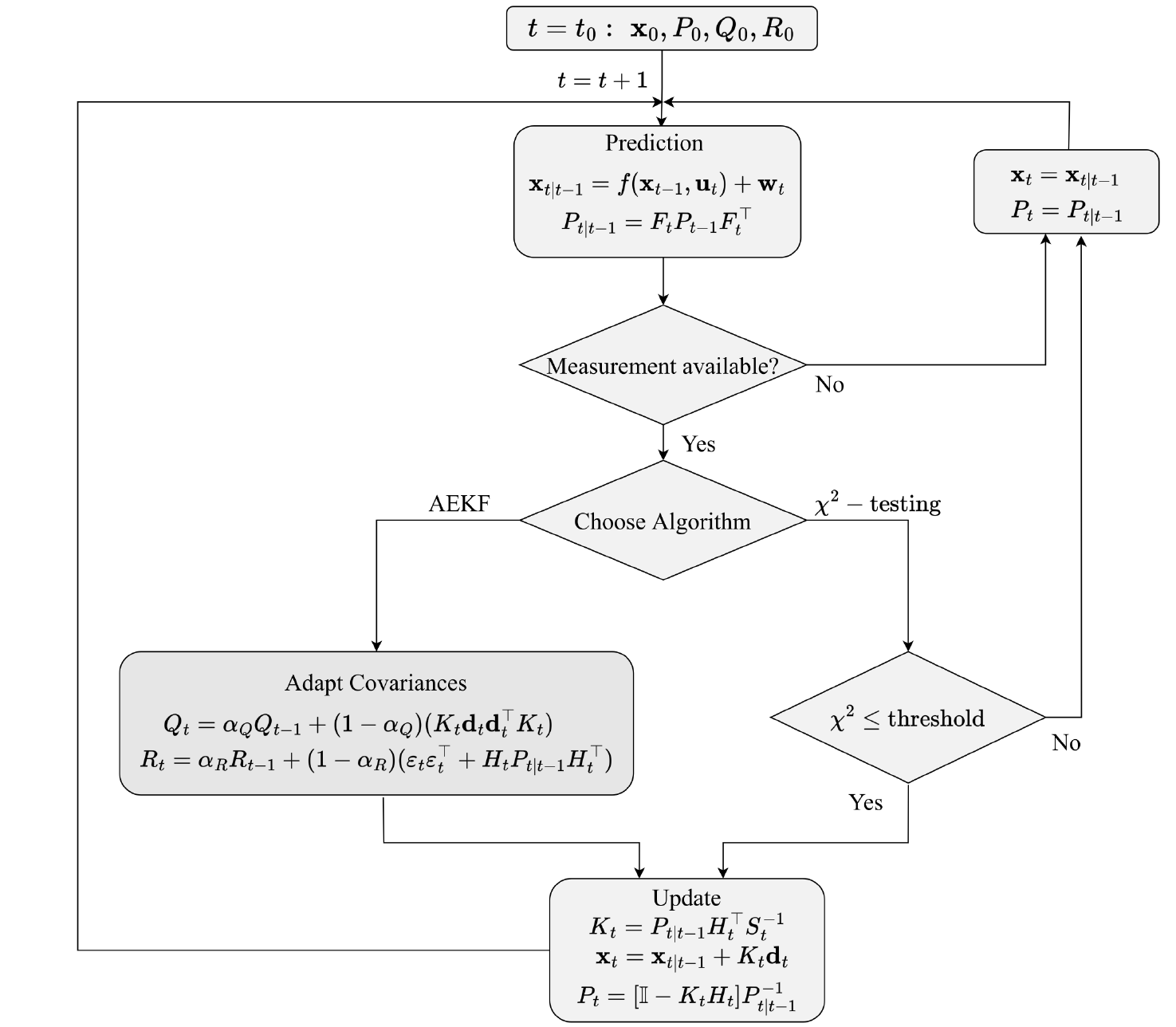}
	\caption{Schematic for Adaptive EKF and EKF with $\chi^2$-testing}
	\label{fig:aekf-chisq}
\end{figure} 
The \textit{correction} step of the EKF is only carried out when, upon receiving a measurement from the ArUco markers, the additional requirement of ${\chi^2 \leq \textit{threshold}}$ is fulfilled, with the value of \textit{threshold} being determined from experiment data. This condition is equivalent to disregarding measurements that are faulty and form outliers. The schematic for the EKF with $\chi^2-$testing is presented in Fig.~\ref{fig:aekf-chisq}. Note that it is possible to use both the AEKF and $\chi^2-$testing in conjunction with each other.



\section{Results and Discussion}
\label{sec:results}
The primary validation for the accuracy of the localization algorithms comes from comparing the $(x, y)-$coordinates of the model car obtained using the EKF, AEKF, and EKF with $\chi^2-$testing respectively with the ground truth obtained by the LiDAR. 
Additionally, in an autonomous driving scenario, the pose determined by the indoor localization algorithm forms an input to the motion planning and control modules of the model car. 
Therefore, we also test the impact of the state estimate obtained from AEKF and EKF with $\chi^2-$testing on the output of the trajectory planning and model predictive controller (MPC) nodes. The detailed implementation of the motion planning and control framework of the model car is beyond the scope of this work and can be found in \cite{folkers2020time} and \cite{rick2019autonomous}. 

\begin{figure}[ht]
	\centering
	\includegraphics[width=0.8\linewidth]{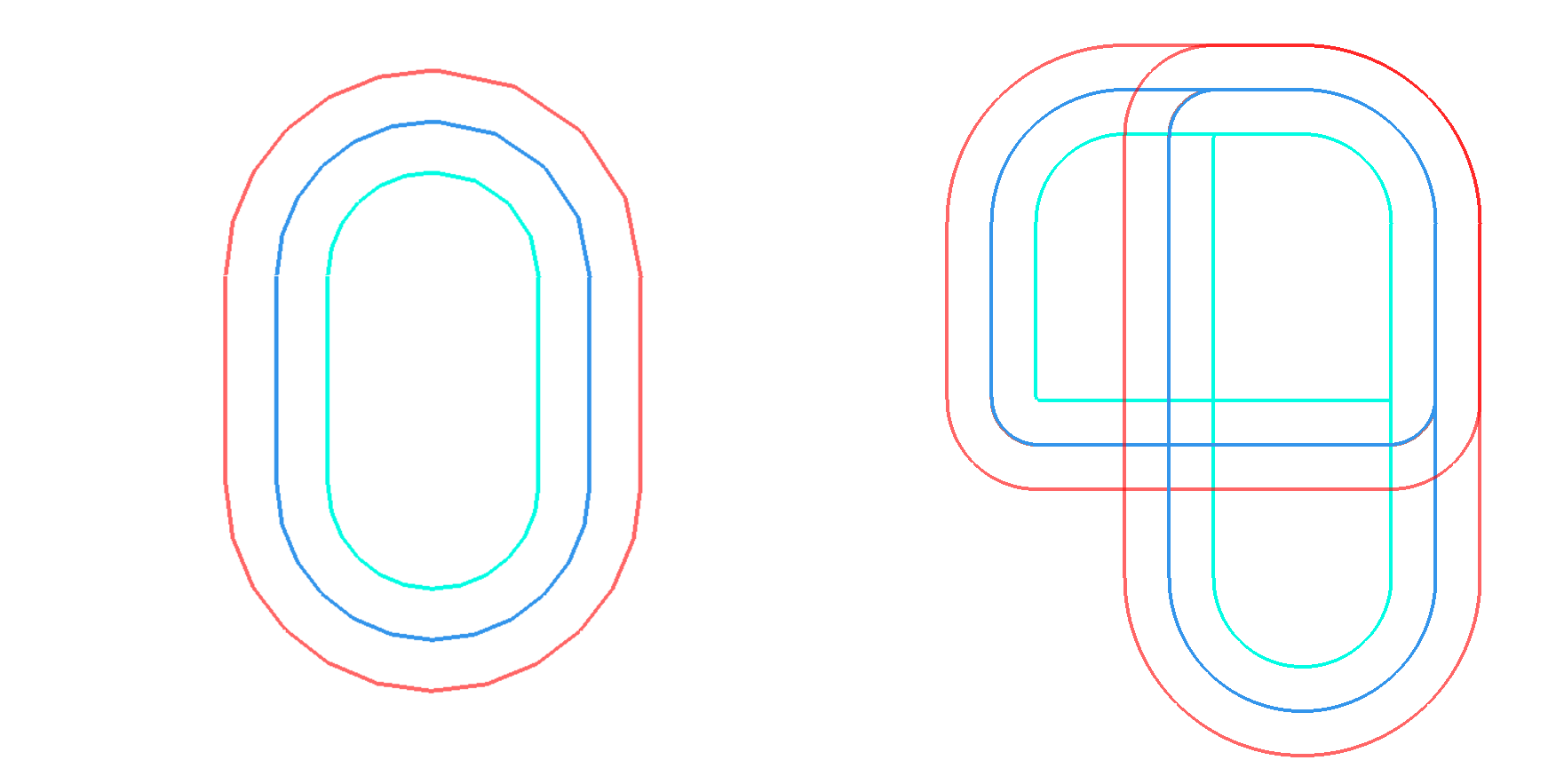}
	\caption{Maps used for real-time validation}
	\label{fig:map_of_validation}
\end{figure}

Fig.~\ref{fig:map_of_validation} shows two different maps which are built for real-time validation of our algorithms. On the left side in Fig.~\ref{fig:map_of_validation}, the \textit{oval map} is presented with the dimensions of 4$\times$6 m. Twelve equispaced ArUco Markers are placed along the outer boundary of the map in order to provide position measurements. Each ArUco marker is assigned with a predefined global position and orientation. A robust validation requires that the conclusions be generalizable to multiple driving scenarios, therefore a second map called the \textit{crossroads map}, shown on the right in Fig.~\ref{fig:map_of_validation}, is also used. The scale of crossroads map is 6$\times$8 m with an intersection. Sixteen equispaced ArUco markers are similarly placed along the map boundary in this case. The results of localization with AEKF and EKF with $\chi^{2}-$testing are discussed in the next subsection, and the EKF approach is taken as the baseline for comparison of algorithm performance. 

\subsection{Filter Initialization and Hyperparameters}
There are several hyperparameters that affect the sensor fusion performance. The process and measurement noise matrices $Q_t$ and $R_t$ appear in all three algorithms. In case of the EKF and EKF with $\chi^2-$testing, the covariance matrices are initialized once and remain constant throughout the filter run, such that $Q_t = Q_{t_0}, R_t = R_{t_0}$. For the AEKF, they are intialized at $t = {t_0}$ and adaptively change depending on the value of $\alpha_Q$ and $\alpha_R$ at each time-step. For consistency, the noise matrices $Q_0$ and $R_0$, as well as the state covariance matrix $P_{t_0}$ are initialized to the same values for all three filters, as shown in Tab.~\ref{tab:hyper_parameters}. For the adaptive tuning of noise matrices in the AEKF, we also need to make a selection of $\alpha_Q$ and $\alpha_R$. This is done by repeating the experiment varying $\alpha_Q$ and $\alpha_R$ from 0 to 1 in steps of 0.1, and finding a combination that minimizes the root mean squared error (RMSE) between the estimated position and the ground truth. For the oval and the crossroads map, the resulting combination of $\{\alpha_Q, \alpha_R\}$ is $\{0.6, 1\}$ and $\{0.9, 1\}$ respectively. Lastly, the value for the $\chi^2$ \textit{threshold} is empirically chosen to be 0.05. 
The sensor fusion algorithms also require an initial estimate for the state vector $\mathbf{x} = [x, y, \varphi, v]^{\top}$. The speed $v$ is trivially initialized to zero, and the rest of the quantities get initialized via the first measurement from the ArUco markers. Consequently, a fiducial marker needs to be in the line of sight of the camera mounted on the model car, otherwise the filters fail to initialize. 
\begin{table}
	\medskip
	\begin{center}
		\caption{Hyperparameters used in presented methods}
		\label{tab:hyper_parameters}
		\begin{tabular}{c c}
			\hline
			Parameters & Values \\
			\hline
			\textit{threshold} & 0.05\\
			$\{\alpha_Q, \alpha_R\}$ Oval & $\{0.6, 1\}$\\
			$\{\alpha_Q, \alpha_R\}$ Crossroads & $\{0.9, 1\}$\\
			$R_{t_0}$ & $\begin{bmatrix} 0.5 & 0.5 & 1  \end{bmatrix}$\\
			$P_{t_0}$ & $\begin{bmatrix} 0.1 & 0.1 & 0.1 & 0.1  \end{bmatrix}$\\
			$Q_{t_0}$ & $\begin{bmatrix} 10^{-3} & 10^{-3} & 3 \times 10^{-4} & 2 \times 10^{-3}  \end{bmatrix}$\\		
			\hline
		\end{tabular}
	\end{center}
\end{table}
\subsection{Localization on Oval and Crossroads Maps}
\begin{figure*}[tp]
	\medskip
	\centering	
	\begin{subfigure}[t]{0.4\linewidth}
		\centering
		\includegraphics[width=\linewidth, trim = 0cm 0.5cm 1.5cm 2.5cm, clip]{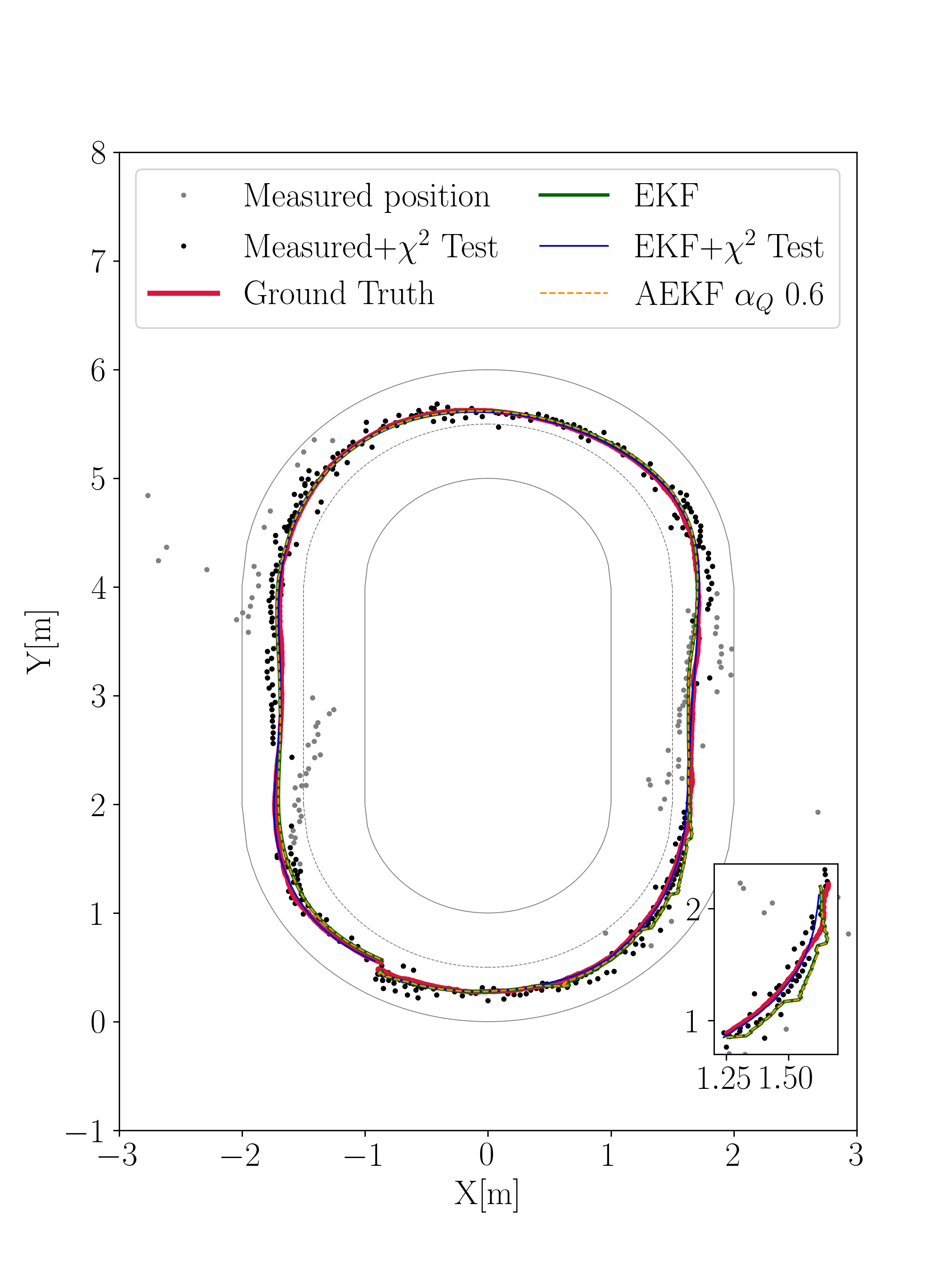}
		\caption{}
		\label{fig:ekf_result_oval}
	\end{subfigure}
	\begin{subfigure}[t]{0.4\linewidth}
		\centering
		\includegraphics[width=\linewidth, trim = 0cm 0.5cm 1.5cm 2.5cm, clip]{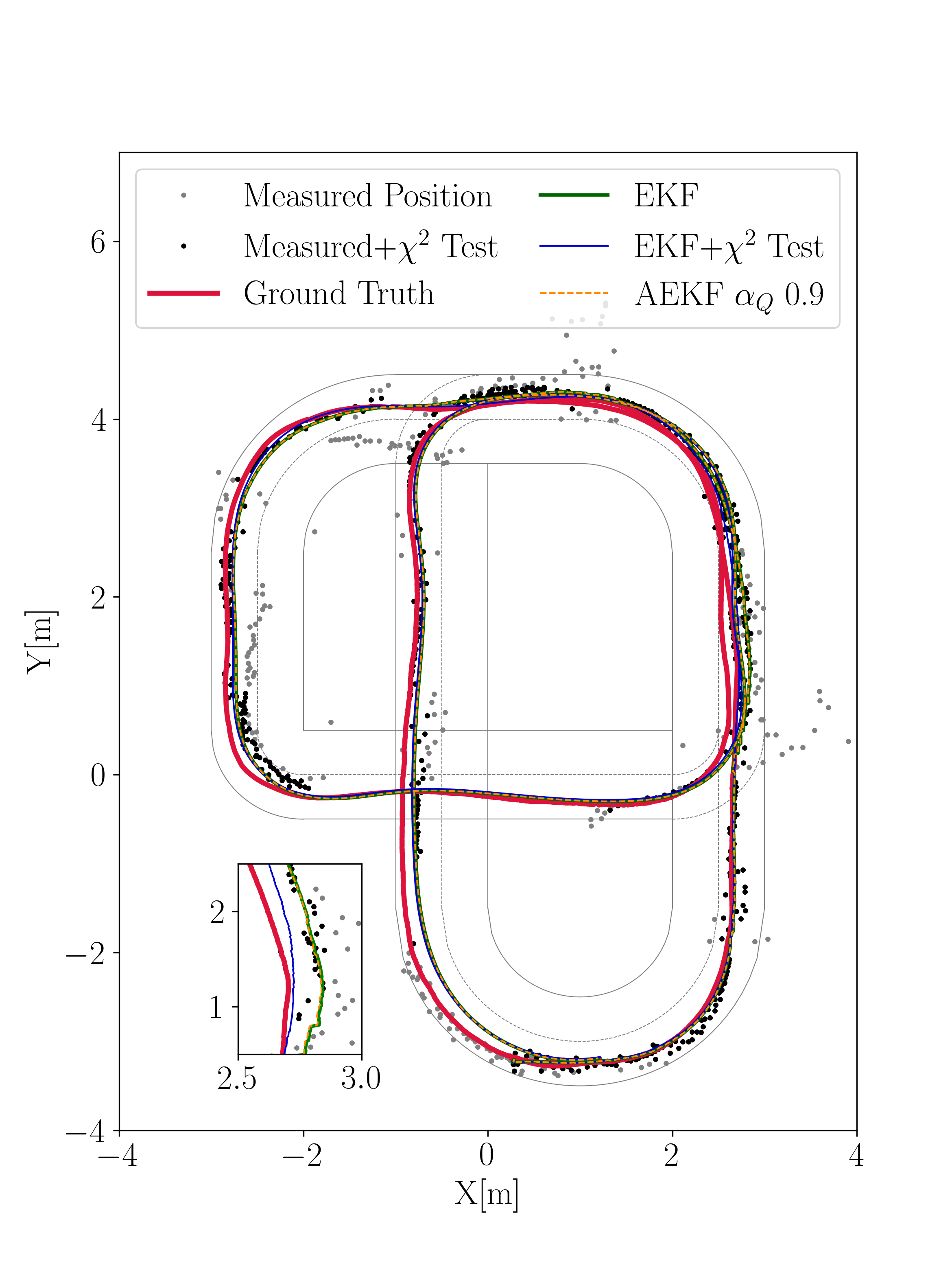}
		\caption{}
		\label{fig:ekf_result_cross_road}
	\end{subfigure}
	\caption{Indoor localization results on (a) the oval map and (b) the crossroads map}
\end{figure*}

In Figs.~\ref{fig:ekf_result_oval} and \ref{fig:ekf_result_cross_road} we plot the positions of the model car obtained from the three sensor-fusion algorithms, overlaid on the oval and crossroads maps. The red line shows the ground truth trajectory that is obtained from the interpolated LiDAR point cloud data, and the positions obtained from the EKF, EKF with $\chi^2-$testing, and AEKF are depicted by the solid dark green line, solid blue line, and the dashed orange line respectively. The gray points show the measured positions that we get from the ArUco markers, and the black points show the same measurements after filtering them via the $\chi^2-$testing method.

\begin{figure}[t!]
	\begin{minipage}{0.45\linewidth}
		\centering
		\includegraphics[width=\linewidth]{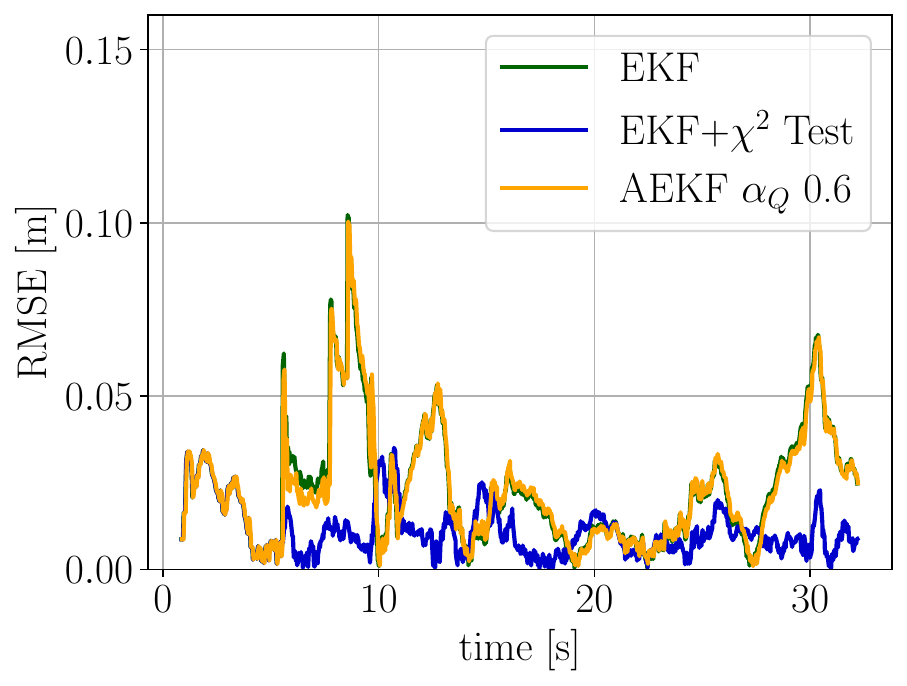}  
		\subcaption{RMS error (oval)}
		\label{fig:rms_oval}
	\end{minipage}
	\begin{minipage}{0.45\linewidth}
		\centering
		\includegraphics[width=\linewidth]{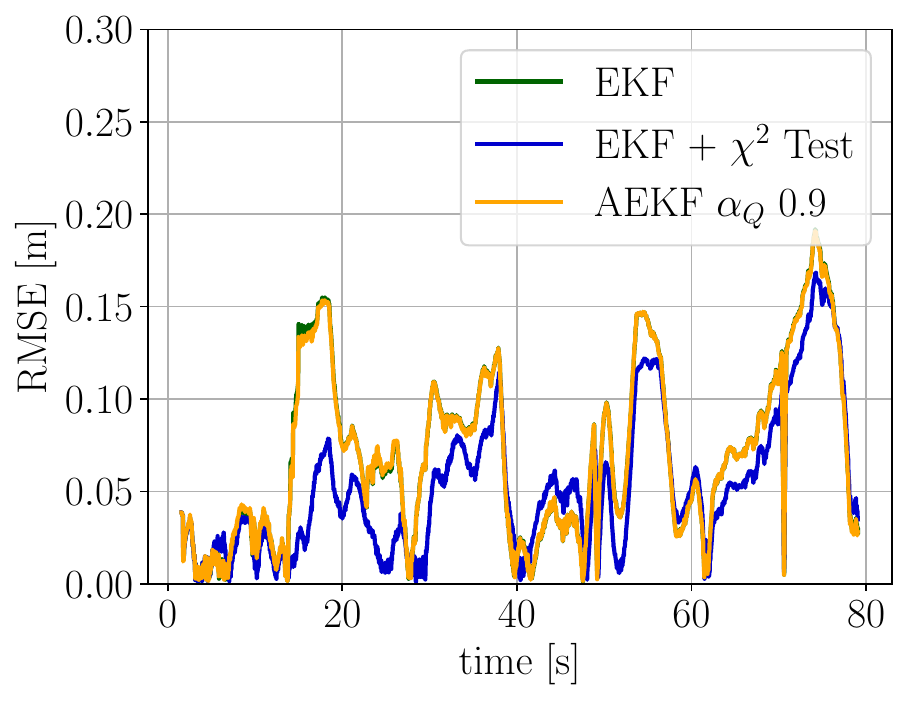}  
		\subcaption{RMS error (crossroads)}
		\label{fig:rms_cross_road}
	\end{minipage}
	\caption{RMSE for the two maps}
	\label{fig:rms_error}
\end{figure}

In general, we observe that the EKF and AEKF have more or less overlapping trajectories, which indicates that the choice of the process noise matrix does not play a huge role in the deviations from the ground truth in the case of our system. During the experiments, it was observed that changing $\alpha_R$ to a value even slightly different from 1.0 leads to severe disturbances in the resulting fused state estimate, which points to the fact that when innovation values grow large due to measurement outliers, they introduce numerical instabilities in the algorithm and lead it to crash. From Fig.~\ref{fig:ekf_result_oval}, we find that deviations and drifts from the ground truth occur the most in regions of the map that contain a significant amount of measurement outliers, as shown in the zoomed-in portion of the oval map. The EKF with $\chi^2-$testing disregards the measurements that are outliers. The resulting trajectory is much closer to the ground truth than the ones for EKF and AEKF. Consequently, we conclude that the choice of a suitable localization algorithm in our system is affected by faulty measurements, rather than an imprecise choice of a noise modeling. This experiment also points out the unreliability of using ArUco markers for positioning alone, as they lead to significant amount of outlier measurements. Similar conclusions can be drawn from the crossroads map in Fig.~\ref{fig:ekf_result_cross_road}, where due to the sharper curves however, a larger drift away from the ground truth can be seen for all three algorithms. The most likely case of that is that the IMU measurements accumulate a gyroscope drift - which directly introduces an IMU integration error into the algorithm. Nevertheless, the EKF with $\chi^2-$testing remains the most accurate algorithm of the three. Conclusions drawn from Figs.~\ref{fig:ekf_result_oval} and \ref{fig:ekf_result_cross_road} are reiterated in Fig.~\ref{fig:rms_error}, which shows the RMSE between the localized positions and the ground truth as a function of time for each of the algorithms, with the RMSE for EKF with $\chi^2-$testing being the smallest for each map. 

\begin{figure}[t]
	\smallskip
	\begin{minipage}{0.45\linewidth}
		\centering
		\includegraphics[height=0.7\linewidth]{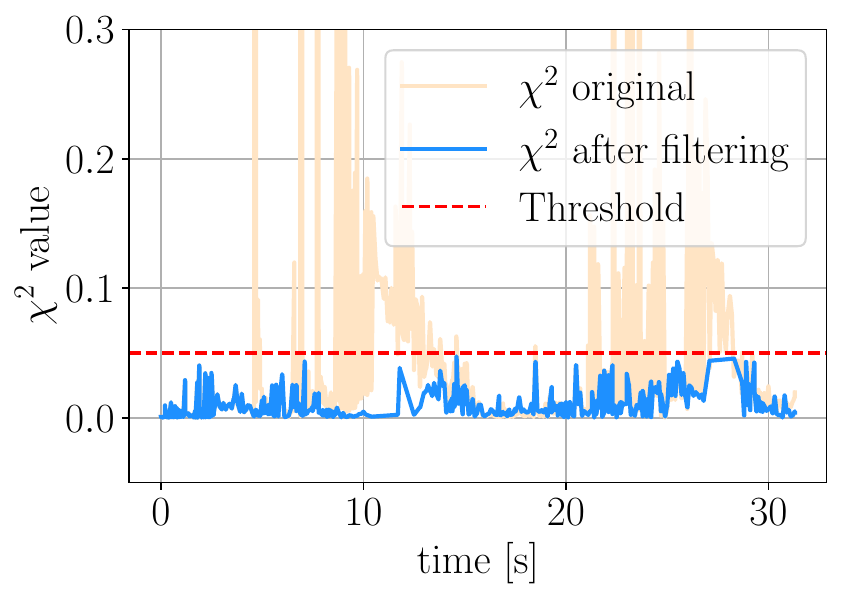}  
		\subcaption{$\chi^{2}$ values vs. time}
		\label{fig:chi_square_test}
	\end{minipage}
	\begin{minipage}{0.45\linewidth}
		\centering
		\includegraphics[height=0.7\linewidth]{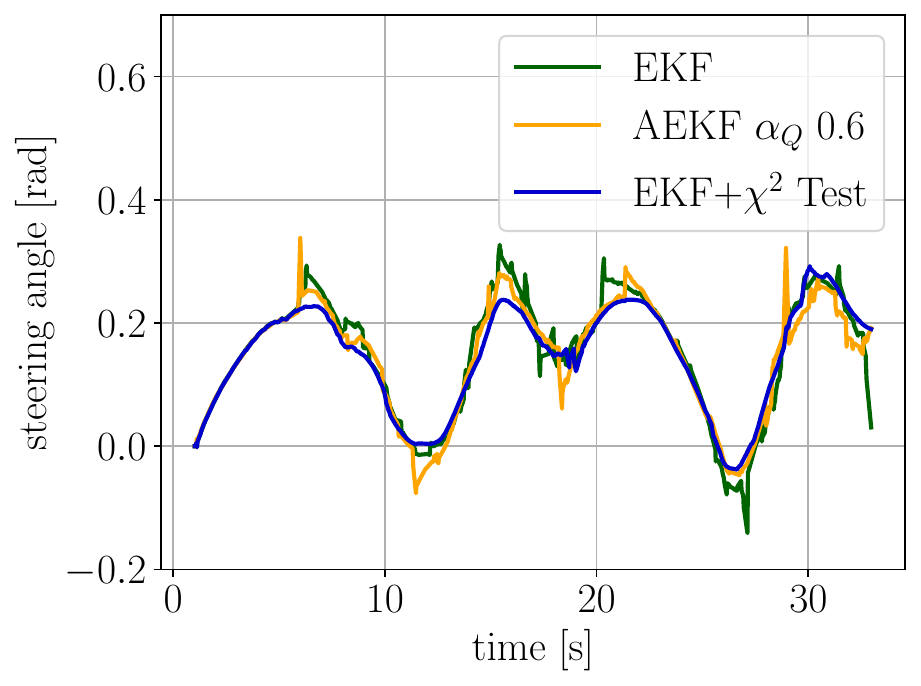}  
		\subcaption{Steering controls vs. time }
		\label{fig:control_signals_oval}
	\end{minipage}
	\caption{The $\chi^{2}$ values and steering controls for the oval map}
\end{figure}

Finally, Fig.~\ref{fig:chi_square_test} shows the $\chi^2$ values before and after filtering for the oval map. This figure is indicative of the importance of choosing a $\chi^2$ threshold judiciously, as too large a value is highly permissive of faulty measurements, whereas too small a value leads to ignoring accurate measurements and an over-reliance on the prediction step of the filter alone. In Fig.~\ref{fig:control_signals_oval}, the steering angle controls obtained from the MPC for each of the three localization algorithms in the oval map are plotted as a function of time. Whereas the faulty measurements lead to sudden jumps in the steering angle for trajectories obtained from EKF and AEKF, the EKF with $\chi^2-$testing leads to a much smoother and jerk-free control output, which is preferable in any driving scenario to reduce machine stress and increase rider comfort. From our experiments, we conclude that the EKF with $\chi^2-$testing leads to more precise indoor localization for our setup. In addition, we also measured the computational time of the EKF with $\chi^2-$testing, which is 0.0519 ms on average and fulfills our real-time requirement.

\section{CONCLUSIONS}
\label{sec:conclusion}
In order to localize the ADAS model car in a pre-built indoor map, we introduced an integrated framework including a ground truth measurement system, pose estimation based on multi-sensor fusion and the related control output. The standard EKF algorithm is extended with adaptive noise tuning and outlier filtering with $\chi^2-$testing. The main contributions of this paper can be summarized as a complete and low-cost indoor localization framework, validation in real-time experiments which present the measurement error of ArUco Marker, and the corresponding outlier filtering method. 






\bibliographystyle{IEEEtran}
\bibliography{references}

\end{document}